\documentclass[conference]{IEEEtran}
\usepackage{algorithmic}
\usepackage{cite}
\usepackage{amsfonts}
\usepackage{amsmath}
\usepackage{amssymb}
\usepackage{booktabs}
\usepackage{caption}
\usepackage{enumitem}
\usepackage{graphicx}
\usepackage[hidelinks]{hyperref}
\usepackage{subcaption}
\usepackage{textcomp}
\usepackage{url}
\usepackage{xcolor}
\def\BibTeX{{\rm B\kern-.05em{\sc i\kern-.025em b}\kern-.08em
T\kern-.1667em\lower.7ex\hbox{E}\kern-.125emX}}
\begin{document}
\title{AutoBnB-RAG: Enhancing Multi-Agent Incident Response with Retrieval-Augmented Generation}
\author{
\IEEEauthorblockN{
Zefang Liu\IEEEauthorrefmark{1}\IEEEauthorrefmark{2}, 
Arman Anwar\IEEEauthorrefmark{2}
}
\IEEEauthorblockA{\IEEEauthorrefmark{1}\textit{Capital One}, San Jose, USA}
\IEEEauthorblockA{\IEEEauthorrefmark{2}\textit{Georgia Institute of Technology}, Atlanta, USA}
\IEEEauthorblockA{liuzefang@gatech.edu}
}
\maketitle
\begin{abstract}
Incident response (IR) requires fast, coordinated, and well-informed decision-making to contain and mitigate cyber threats. While large language models (LLMs) have shown promise as autonomous agents in simulated IR settings, their reasoning is often limited by a lack of access to external knowledge. In this work, we present AutoBnB-RAG, an extension of the AutoBnB framework that incorporates retrieval-augmented generation (RAG) into multi-agent incident response simulations. Built on the Backdoors \& Breaches (B\&B) tabletop game environment, AutoBnB-RAG enables agents to issue retrieval queries and incorporate external evidence during collaborative investigations. We introduce two retrieval settings: one grounded in curated technical documentation (RAG-Wiki), and another using narrative-style incident reports (RAG-News). We evaluate performance across eight team structures, including newly introduced argumentative configurations designed to promote critical reasoning. To validate practical utility, we also simulate real-world cyber incidents based on public breach reports, demonstrating AutoBnB-RAG's ability to reconstruct complex multi-stage attacks. Our results show that retrieval augmentation improves decision quality and success rates across diverse organizational models. This work demonstrates the value of integrating retrieval mechanisms into LLM-based multi-agent systems for cybersecurity decision-making.
\end{abstract}
\begin{IEEEkeywords}
incident response, large language models, multi-agent systems, retrieval-augmented generation, cybersecurity
\end{IEEEkeywords}
\section{Introduction}

As cyber threats become more frequent, sophisticated, and multi-phased, effective incident response (IR)~\cite{west1998handbook,mandia2001incident,kruse2001computer,ahmad2012incident,luttgens2014incident} has become a critical capability for organizational resilience. Incident response demands timely decision-making, coordination across specialized roles, and the ability to adapt to incomplete and evolving information. Traditional approaches rely heavily on human teams, structured protocols, and expert judgment, which can be slow or inconsistent under operational stress. To address these limitations, recent research has explored the use of large language models (LLMs) as autonomous agents capable of supporting or simulating incident response teams.

LLMs have demonstrated strong performance in multi-agent collaboration tasks~\cite{naveed2023comprehensive,guo2024large} due to their capabilities in natural language understanding, planning, and communication. These capabilities have enabled progress in domains such as scientific reasoning~\cite{ni2024mechagents}, healthcare decision-making~\cite{wang2024colacare}, economic retrieval~\cite{liu2025econwebarena}, supply chain management~\cite{quan2024invagent}, and customer relationship~\cite{quan2025crmagent}. In cybersecurity, LLMs have been evaluated on benchmarks~\cite{liu2024review,liu2023secqa,liu2024cyberbench,tihanyi2024cybermetric} and deployed in simulation environments to emulate defenders or analysts~\cite{motlagh2024large,xu2024large,hays2024employing,caviglione2024dawn}. One prominent framework for such simulations is Backdoors \& Breaches (B\&B)~\cite{backdoorsandbreaches,young2021backdoors,seiler2025improving}, a structured tabletop game designed to model realistic IR scenarios. AutoBnB~\cite{liu2024multi,liu2025autobnb} extended this framework by enabling LLM-based agents to collaborate through structured dialogue in uncovering attack sequences under various team structures.

\begin{figure}[t!]
    \centering
    \includegraphics[width=\linewidth]{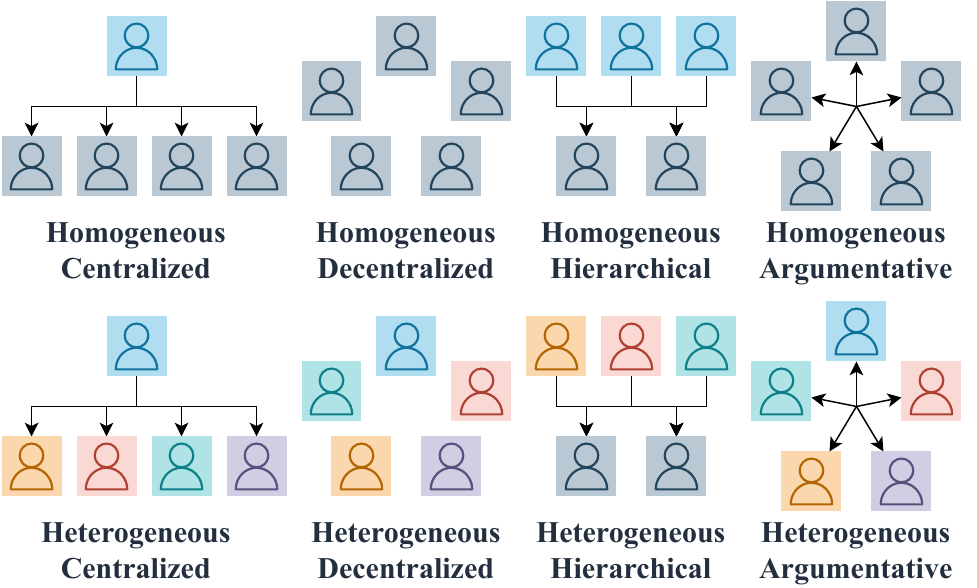}
    \caption{Team structures evaluated in LLM-driven incident response simulations using the \textit{Backdoors \& Breaches} framework.}
    \label{fig:team-structures}
\end{figure}

This paper introduces AutoBnB-RAG\footnote{\url{https://github.com/zefang-liu/AutoBnB}}, an extension of the AutoBnB framework~\cite{liu2024multi,liu2025autobnb} that equips LLM agents with retrieval capabilities during simulated incident response. Although LLMs are effective in reasoning and dialogue, they can suffer from hallucinations or gaps in factual knowledge, especially when faced with domain-specific or evolving threats. AutoBnB-RAG builds on the concept of retrieval-augmented generation (RAG)~\cite{lewis2020retrieval,gao2023retrieval}, allowing agents to issue queries and incorporate external knowledge dynamically throughout the simulation. We define two retrieval settings: RAG-Wiki, which provides access to curated technical documentation, and RAG-News, which offers narrative-style incident response stories. We evaluate eight team structures, including two newly introduced argumentative configurations that promote internal critique and reflective reasoning. To validate the framework beyond synthetic settings, we also simulate three real-world cybersecurity incidents based on public breach disclosures. This integration of retrieval and multi-agent coordination offers a more grounded and adaptive framework for simulating LLM-driven incident response.

\section{Related Work}

Recent research has explored the integration of large language models (LLMs) with retrieval-augmented generation (RAG) to support cybersecurity operations. GenDFIR~\cite{loumachi2025advancing} demonstrated the potential of zero-shot LLMs combined with RAG for forensic timeline reconstruction. CyberRAG~\cite{blefari2025cyberrag} introduced an agentic framework that combines iterative retrieval with specialized classifiers to enhance cyber-attack classification and explanation. MoRSE~\cite{simoni2025morse} employed parallel RAG pipelines over heterogeneous sources to improve QA accuracy in cybersecurity contexts. Graph-enhanced approaches such as CyKG-RAG~\cite{kurniawan2024cykg} and GraphCyRAG~\cite{rahman2024retrieval} integrated structured knowledge graphs into the RAG pipeline, improving contextual grounding and retrieval precision. Other domain-specific applications include cyber-attack attribution~\cite{rajapaksha2024rag}, cybersecurity education~\cite{zhao2024ontology}, and threat tracing using graph-based RAG modeling~\cite{jeon2024rag}. Our work, AutoBnB-RAG, advances this line of research by embedding RAG into a multi-agent simulation framework for incident response, enabling LLM agents to dynamically retrieve and share external knowledge during collaborative decision-making.

\section{Methodology}

In this section, we present the simulation framework, team structures, and retrieval-augmented generation setup used to evaluate multi-agent incident response with large language models.

\subsection{Simulation Framework}

\begin{figure}[h!]
    \centering
    \begin{subfigure}[b]{0.48\columnwidth}
        \centering
        \includegraphics[width=\textwidth]{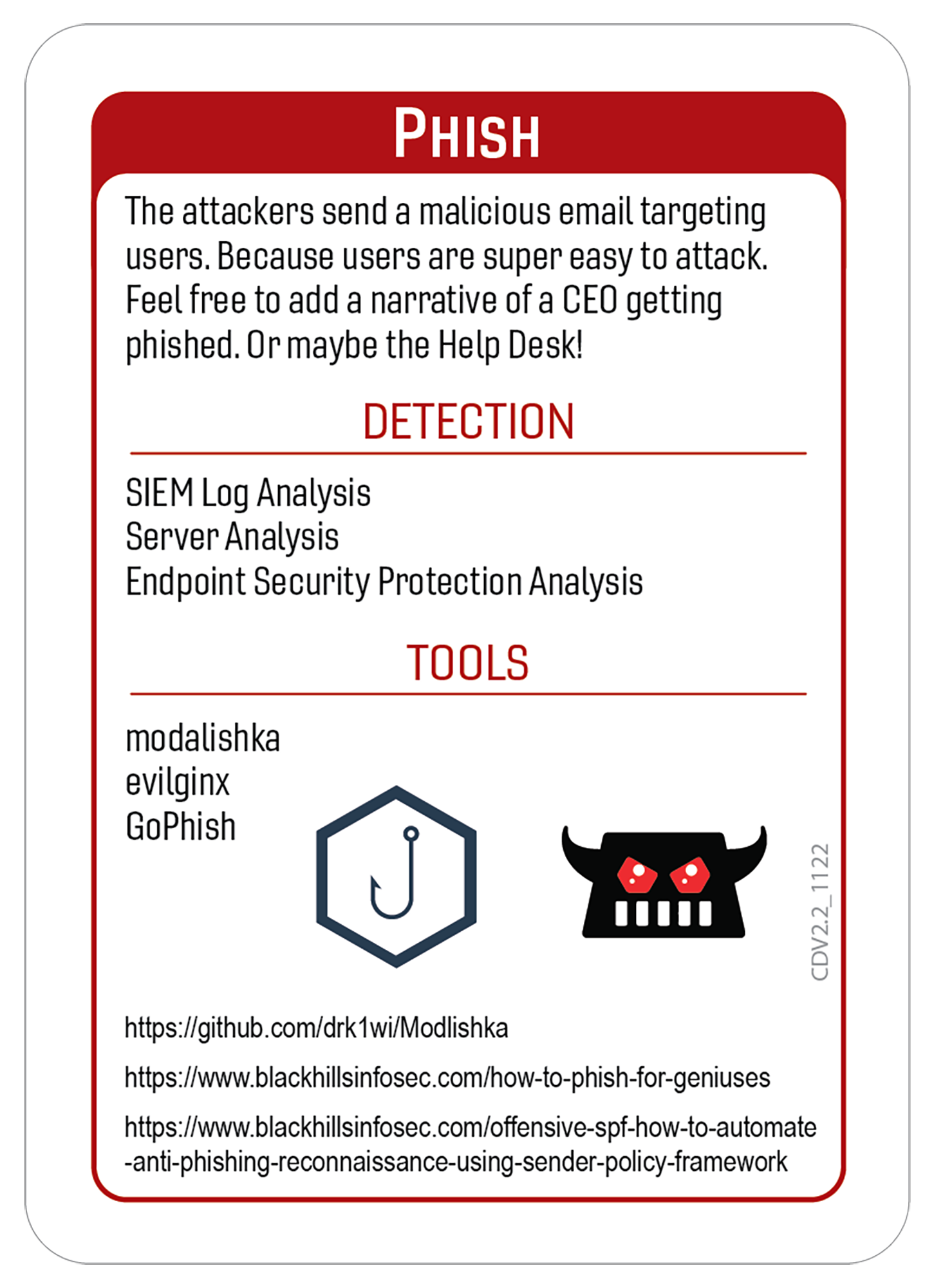}
        \caption{Initial Compromise card}
    \end{subfigure}
    \begin{subfigure}[b]{0.48\columnwidth}
        \centering
        \includegraphics[width=\textwidth]{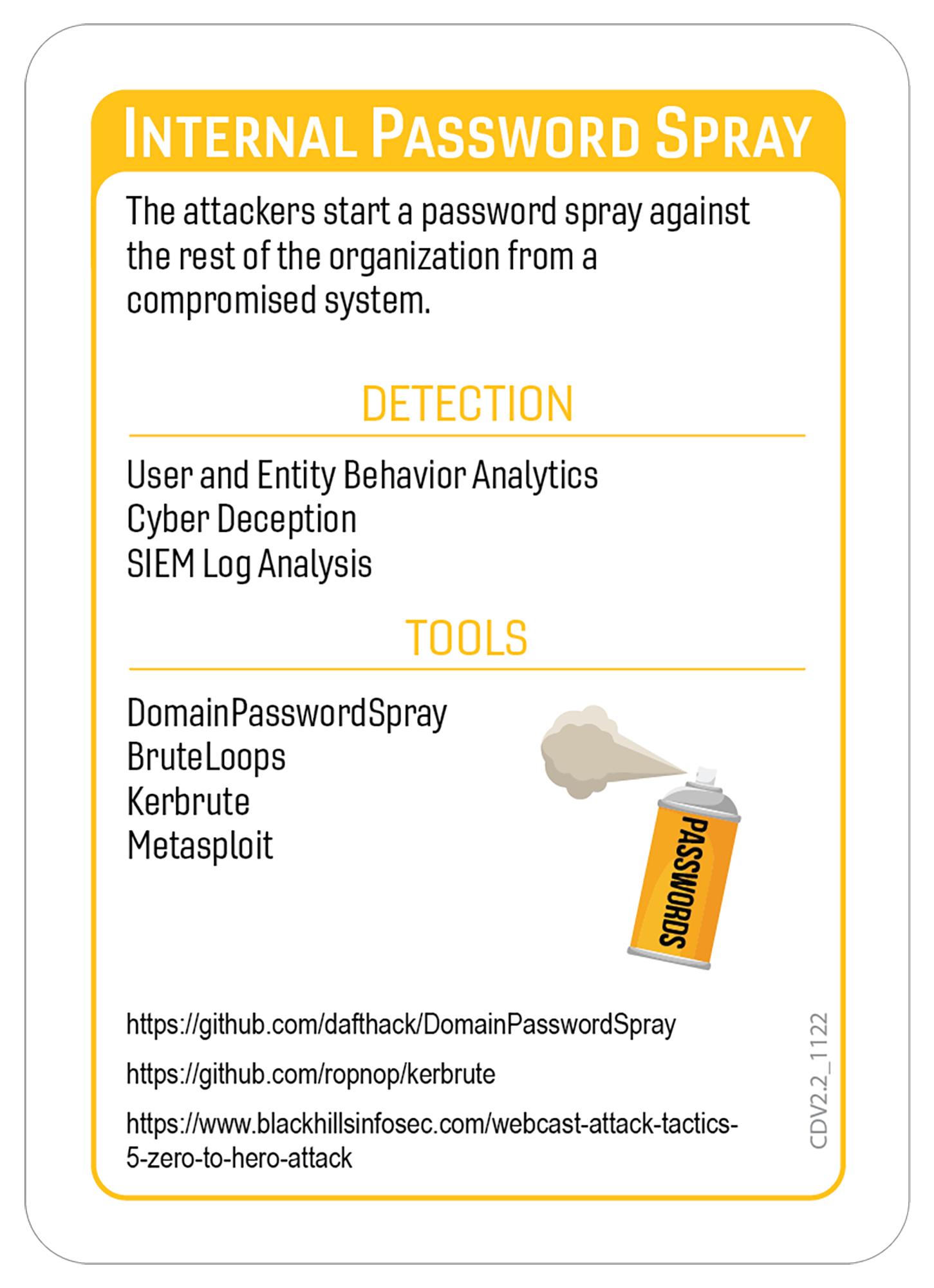}
        \caption{Pivot and Escalate card}
    \end{subfigure}
    \begin{subfigure}[b]{0.48\columnwidth}
        \centering
        \includegraphics[width=\textwidth]{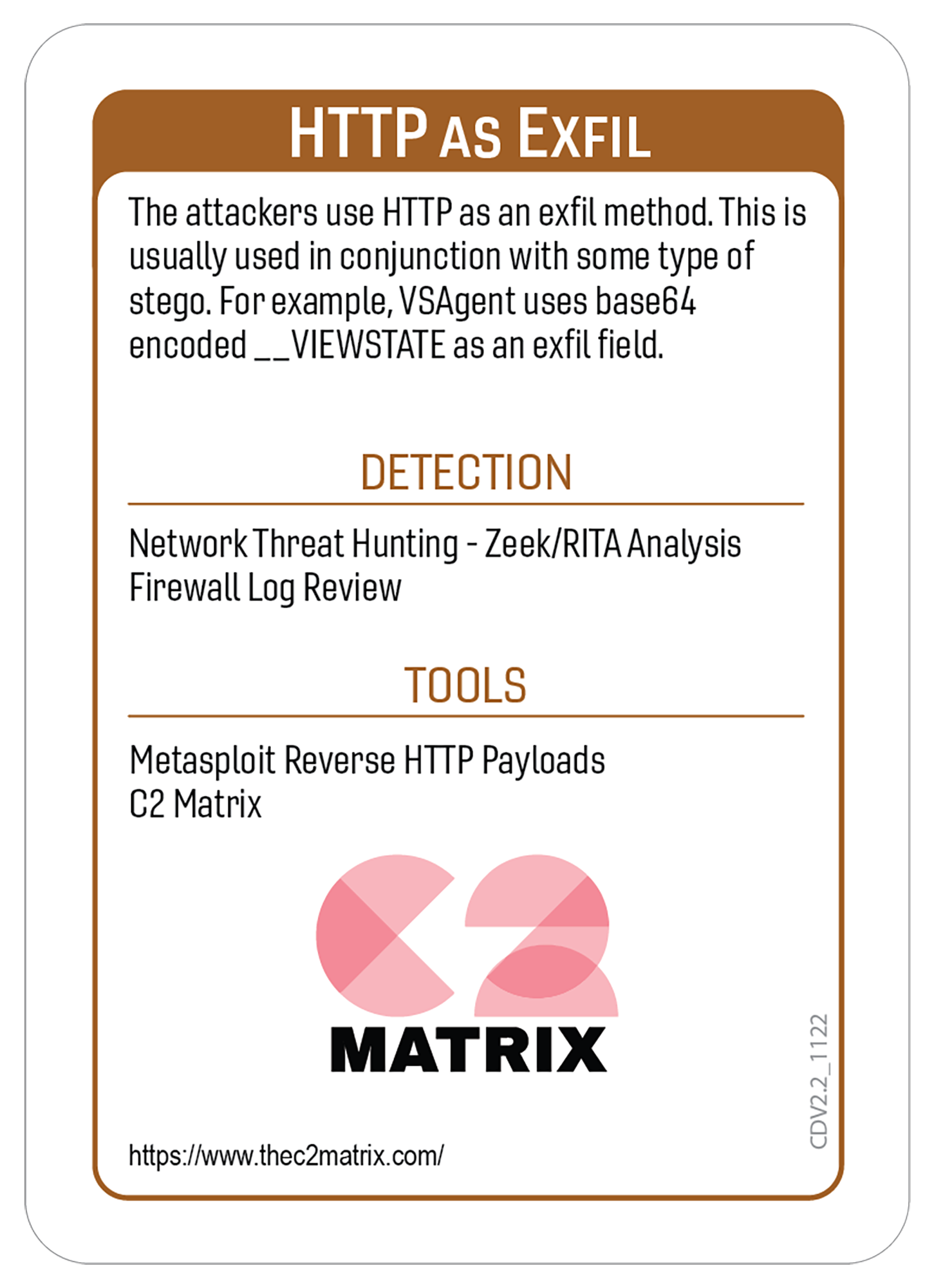}
        \caption{C2 and Exfil card}
    \end{subfigure}
    \begin{subfigure}[b]{0.48\columnwidth}
        \centering
        \includegraphics[width=\textwidth]{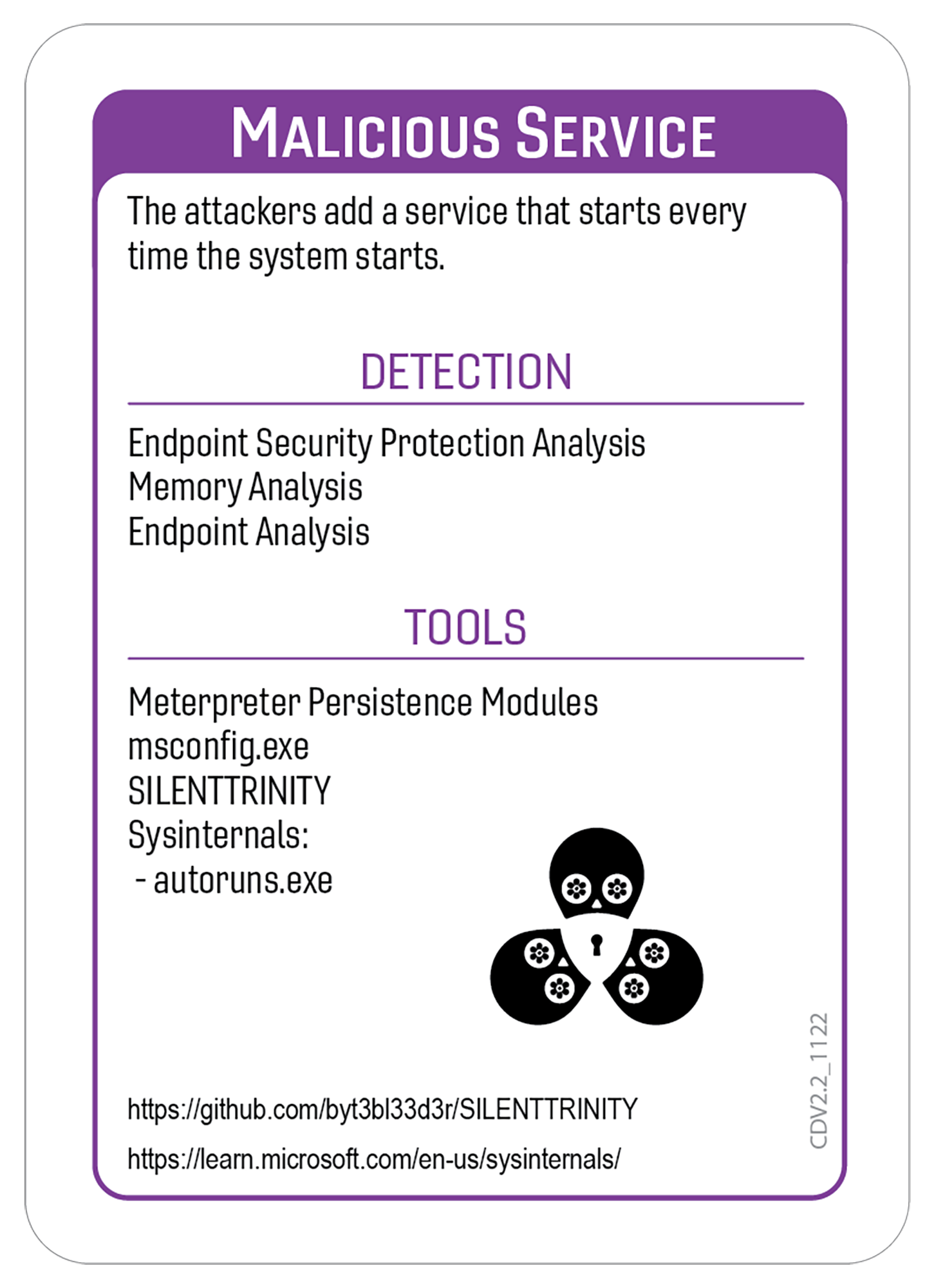}
        \caption{Persistence card}
    \end{subfigure}
    \begin{subfigure}[b]{0.48\columnwidth}
        \centering
        \includegraphics[width=\textwidth]{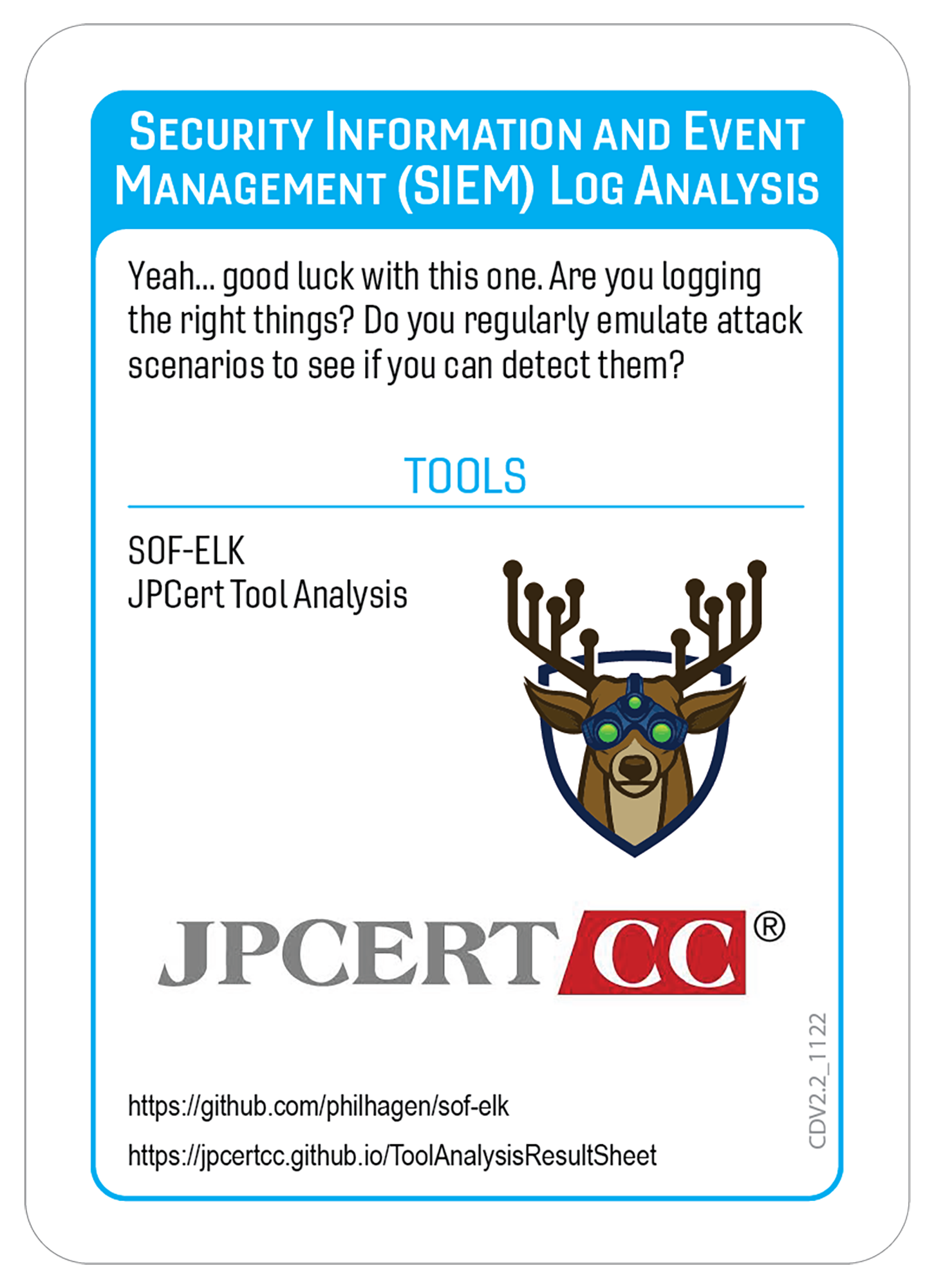}
        \caption{Procedure card}
    \end{subfigure}
    \begin{subfigure}[b]{0.48\columnwidth}
        \centering
        \includegraphics[width=\textwidth]{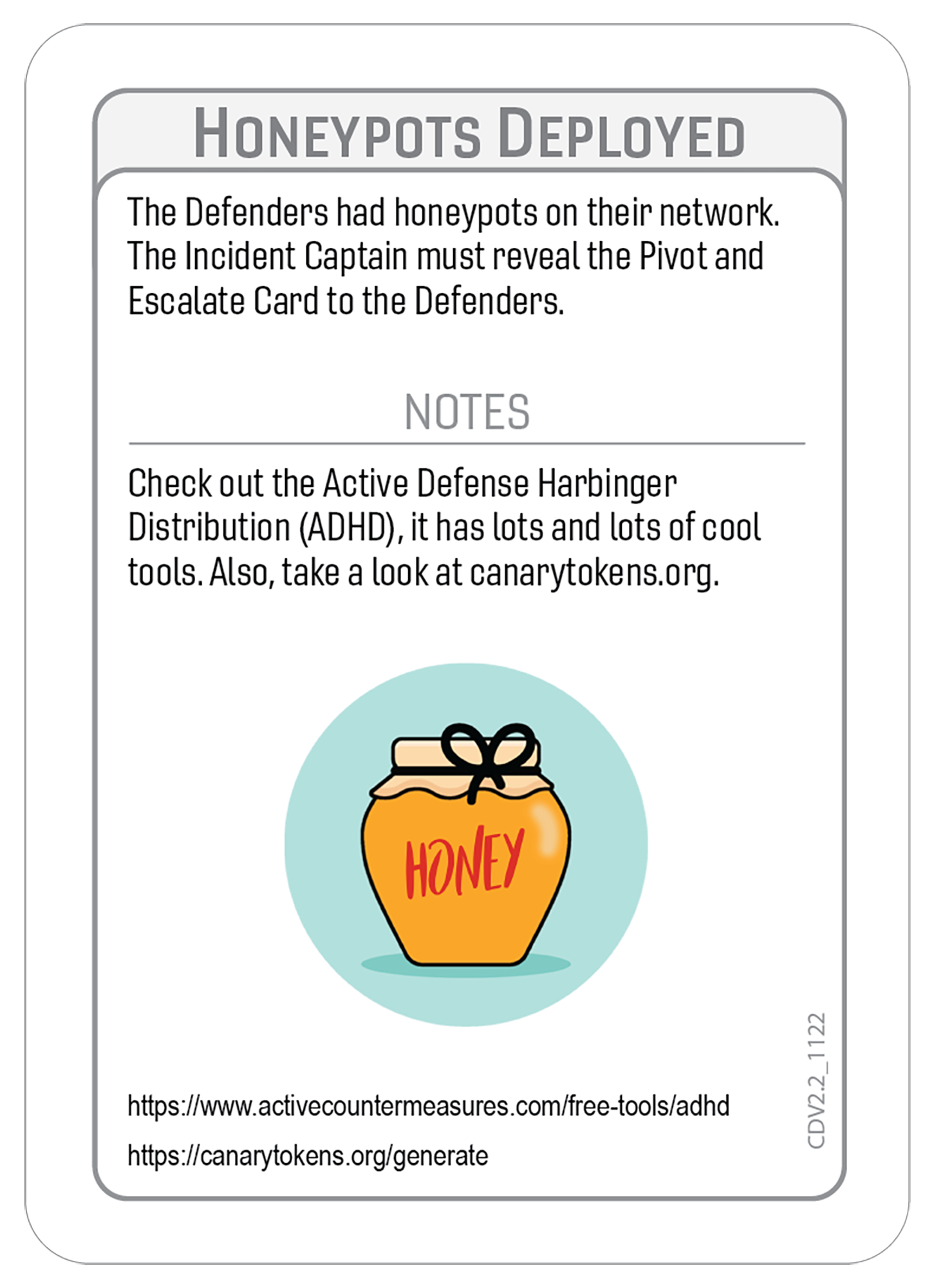}
        \caption{Inject card}
    \end{subfigure}
    \caption{Examples of \textit{Backdoors \& Breaches} cards used in this study. Image source: \textit{Black Hills Information Security}.}
    \label{fig:bnb-cards}
\end{figure}

We base our simulation on Backdoors \& Breaches\footnote{\url{https://www.blackhillsinfosec.com/tools/backdoorsandbreaches/}} (B\&B)~\cite{backdoorsandbreaches}, a cooperative cybersecurity card game designed to emulate real-world incident response scenarios. The game centers around a structured challenge in which a defending team must uncover a sequence of four hidden attack stages. These stages include initial compromise, pivot and escalate, command and control (C2) with exfiltration, and persistence. The full card set includes over 50 unique cards, organized into 13 initial compromise cards, 12 pivot and escalate cards, 7 C2 and exfiltration cards, and 14 persistence cards. These are complemented by 12 procedure cards that represent common detection or investigative techniques. A detailed listing of all card types and categories is provided in Appendix~\ref{appendix:card-list}. While B\&B also includes inject and consultant cards for added variability, we omit these elements in our simulation to maintain a controlled evaluation setup.

Gameplay begins with one agent assuming the role of the incident captain, who randomly selects one attack card from each of the four attack categories to define the hidden scenario. The defending agents are given access to a pool of procedure cards, with four of them marked as established procedures, which provide a +3 modifier on dice rolls due to their perceived reliability. Each turn, the defenders collaboratively choose a single procedure card and roll a 20-sided die to determine whether the attempt is successful. A roll of 11 or higher reveals a hidden attack card if the selected procedure is relevant to it. The team wins if all four attack cards are revealed within 10 turns; otherwise, the game ends in failure.

Following AutoBnB~\cite{liu2024multi,liu2025autobnb}, to operationalize B\&B for systematic experimentation, we implement it as a multi-agent simulation environment. Human players are replaced by large language model (LLM)-based agents that communicate, reason, and act within the bounds of the game's rules. The simulation environment automates the mechanics of card management, dice rolling, and game progression. Each scenario consists of one incident captain and five defender agents whose roles, expertise, and communication strategies vary based on predefined team structures. This setup enables consistent and repeatable evaluation of different organizational configurations and allows us to investigate how retrieval-augmented generation influences multi-agent incident response under realistic constraints.

\subsection{Team Structures}

We evaluate eight team structures that model different organizational approaches to multi-agent incident response, as illustrated in Figure~\ref{fig:team-structures}. The original six configurations from AutoBnB~\cite{liu2024multi,liu2025autobnb} vary along two dimensions: leadership and expertise. Centralized teams are directed by a designated leader, while decentralized teams rely on collective decision-making. Hierarchical teams introduce mixed experience levels, where senior agents guide others. Each of these can be homogeneous, with all agents acting as generalists, or heterogeneous, with members assigned specific domain expertise such as endpoint security, log analysis, or threat detection.

To expand this design space, we introduce two new structures: homogeneous argumentative and heterogeneous argumentative. These teams include agents that adopt an explicitly critical stance, challenging peer proposals and offering alternative perspectives during collaborative planning. The argumentative role is intended to stimulate deeper analysis and reduce groupthink. In the homogeneous version, all agents are generalists engaging in argumentative reasoning. In the heterogeneous version, this behavior is embedded within a mix of domain-specialized agents. Together, these eight configurations allow us to examine how team composition and reasoning style affect incident response effectiveness.

\subsection{Retrieval-Augmented Generation}

\begin{figure}[h!]
    \centering
    \includegraphics[width=.85\linewidth]{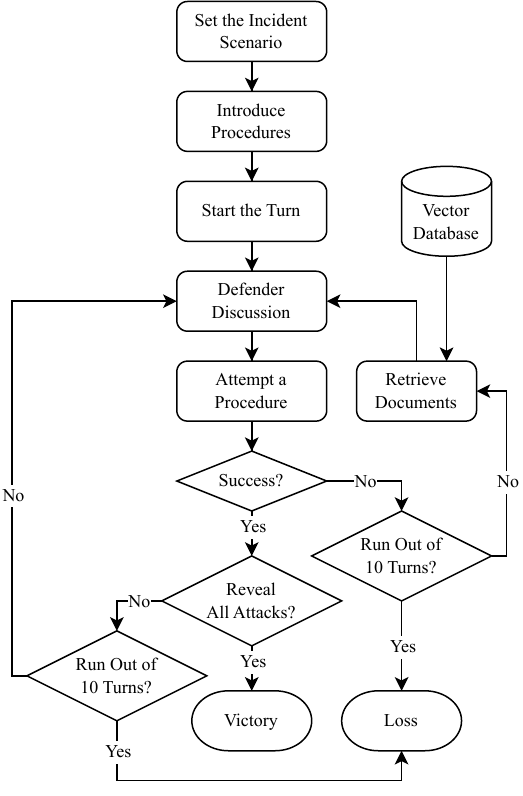}
    \caption{Gameplay flow of AutoBnB-RAG, illustrating the interaction loop between defenders, retrieval, and success conditions.}
    \label{fig:autobnb-rag}
\end{figure}

To enhance reasoning and contextual understanding during gameplay, we integrate a retrieval-augmented generation (RAG) mechanism into the simulation pipeline. This functionality is introduced as a post-attempt step in the turn sequence, specifically after a procedure attempt has been resolved. Following a  failed roll, the incident captain initiates a retrieval operation to surface relevant external information that can assist the defenders. This capability simulates the real-world practice of cybersecurity teams consulting documentation, threat intelligence reports, or knowledge bases when facing uncertainty or investigative dead ends. The full gameplay loop, including the RAG interaction points, is illustrated in Figure~\ref{fig:autobnb-rag}.

A dedicated retrieval agent is added to the agent environment. This agent does not participate in discussion or reasoning but is responsible for handling all retrieval function calls. It receives concise queries from the incident captain and silently returns the relevant results, which are then shared with the group. The incident captain's responsibilities are updated to include identifying when retrieval is appropriate, constructing a meaningful query, and relaying the retrieved knowledge to the defenders. To encourage its use, the detection-checking logic is also modified: on a failed procedure attempt, the system prompts the incident captain to issue a retrieval query using relevant scenario keywords.

The retrieval process is fully integrated into the group chat structure. The retrieval agent is included in the communication graph, and the group chat manager ensures proper speaker transitions. This seamless integration allows the team to access external cybersecurity knowledge in context, without disrupting the natural flow of the game. To explore different retrieval styles, we define two settings: RAG-Wiki, which retrieves from a curated collection of technical articles and documentation, and RAG-News, which retrieves from a synthetic corpus of narrative-style incident reports. These two settings provide contrasting forms of external context, with one grounded in factual reference material and the other in realistic storytelling. We describe the construction of each knowledge source in the following subsections.

\subsection{Webpage Collection}

For the RAG-Wiki setting, we enhanced the AutoBnB framework with retrieval-augmented generation by integrating a curated set of 125 webpages containing relevant cybersecurity knowledge. These webpages were collected from sources such as Wikipedia, Microsoft Learn, MITRE ATT\&CK, OWASP, and leading cybersecurity blogs, with their overall distribution summarized in Table~\ref{tab:webpage-distribution}. The selected documents cover technical explanations, threat models, and practical guidance related to the attack and procedure cards used in the Backdoors \& Breaches simulation. Topics include access token manipulation, ARP spoofing, DLL injection, phishing, insider threats, malware injection, and defensive strategies such as SIEM analysis, deception technology, and endpoint detection. By grounding agent reasoning and discussions in this knowledge base, we aimed to provide contextual clarity and factual support for each decision made during simulated incident response. This RAG integration allows defender agents to retrieve and incorporate real-world cybersecurity insights into their collaborative actions.

\begin{table}[h]
\centering
\small
\caption{Distribution of webpages collected for the RAG-Wiki setting.}
\label{tab:webpage-distribution}
\begin{tabular}{lrr}
\toprule
\textbf{Source Category}        & \textbf{Count} & \textbf{Percentage} \\
\midrule
Wikipedia                       & 67   & 53.6\% \\
MITRE ATT\&CK                   & 9    & 7.2\%  \\
Microsoft Learn / Support       & 6    & 4.8\%  \\
CISA / Government               & 3    & 2.4\%  \\
Cybersecurity Blogs / Vendors   & 27   & 21.6\% \\
Other                           & 13   & 10.4\% \\
\midrule
\textbf{Total}                  & 125  & 100\% \\
\bottomrule
\end{tabular}
\end{table}

\subsection{News Generation}

For the RAG-News setting, we generated 100 synthetic news-style incident reports to serve as retrieval-augmented knowledge grounded in realistic narrative form. These stories were produced using a structured prompt template in Appendix~\ref{prompt-template-news} designed to simulate plausible multi-stage cyberattacks, inspired by the Backdoors \& Breaches card game. To ensure reliability, we additionally conducted manual checks on some sampled stories to validate their narrative quality and adherence to realistic investigative logic. Each news begins with an original title and follows a fictional internal cybersecurity team as they investigate and respond to an unfolding incident. The attack path, comprising stages like initial compromise, privilege escalation, persistence, and data exfiltration, is gradually revealed through the team's application of various procedure cards. The narratives incorporate both successful and failed investigative efforts, highlighting how different procedures either uncovered or missed critical attack vectors. Importantly, each scenario is written to mirror the logic and uncertainty of real-world incident response, offering defenders contextualized examples of how investigations might unfold. To promote coverage and diversity, we use different combinations of attack and procedure cards for news generation and the downstream AutoBnB-RAG evaluation, ensuring minimal overlap and broader generalization across retrieved content.

\section{Experiments}

To evaluate the capabilities of AutoBnB-RAG, we design simulations that test its performance across varied team structures, retrieval settings, and attack scenarios.

\subsection{Experimental Setup}

We follow the simulation protocol from the original AutoBnB framework, using the AutoGen~\cite{wu2023autogen} system with GPT-4o~\cite{achiam2023gpt} as the base model and a temperature setting of 0.7. Each team consists of five defender agents assigned roles based on one of eight predefined structures. The six original structures include: \textbf{homogeneous centralized} (1 team leader and 4 generalist members), \textbf{heterogeneous centralized} (1 leader and 4 domain experts), \textbf{homogeneous decentralized} (5 generalists), \textbf{heterogeneous decentralized} (5 domain experts), \textbf{homogeneous hierarchical} (3 generalist experts and 2 beginners), and \textbf{heterogeneous hierarchical} (3 domain experts and 2 beginners). We introduce two additional team structures to examine the impact of structured disagreement: \textbf{homogeneous argumentative} and \textbf{heterogeneous argumentative}, created by modifying the decentralized versions to assign argumentative roles to all team members. These agents contribute as generalists or experts while actively promoting critical discussion by questioning peer suggestions and offering alternative reasoning. Each team structure is evaluated over 30 independent simulation runs to ensure consistency and statistical robustness.

For retrieval-augmented settings, we use a default configuration that retrieves the top 3 most relevant documents per query. Documents are split into overlapping chunks of 5,000 characters with 500 characters of overlap using a recursive character-based text splitting strategy provided by LangChain~\cite{langchain}. Retrieved passages are stored and indexed using Chroma~\cite{chroma} as the vector database backend with text-embedding-3-small~\cite{openai2024embeddings}. This setup ensures that agents receive contextually relevant and sufficiently detailed information to support their decision-making throughout the simulation.

\subsection{Simulation Example}

\begin{table*}[h!]
    \centering
    \small
    \caption{Turn-by-turn game trajectory from a simulation using the homogeneous centralized team structure with RAG-News.}
    \label{tab:game-trajectory}
    \begin{tabular}{clccclc}
        \toprule
        \textbf{Turn} & \textbf{Procedure} & \textbf{Roll} & \textbf{Modifier} & \textbf{Success} & \textbf{Revealed Incident} & \textbf{Retrieval} \\
        \midrule
        1  & Endpoint Analysis                            & 17 & +3 & Yes & Local Privilege Escalation & No \\
        2  & User and Entity Behavior Analytics           & 10 & +3 & No  & -                          & Yes \\
        3  & Network Threat Hunting - Zeek/RITA Analysis  & 5  & +0 & No  & -                          & Yes \\
        4  & Firewall Log Review                          & 4  & +0 & No  & -                          & Yes \\
        5  & Endpoint Security Protection Analysis        & 20 & +0 & Yes & Application Shimming       & No \\
        6  & Network Threat Hunting - Zeek/RITA Analysis  & 18 & +0 & Yes & Social Engineering         & No \\
        7  & SIEM Log Analysis                            & 10 & +0 & No  & -                          & Yes \\
        8  & Network Threat Hunting - Zeek/RITA Analysis  & 4  & +0 & No  & -                          & Yes \\
        9  & User and Entity Behavior Analytics           & 3  & +3 & No  & -                          & Yes \\
        10 & Network Threat Hunting - Zeek/RITA Analysis  & 11 & +0 & Yes & HTTP as Exfil              & No \\
        \bottomrule
    \end{tabular}
\end{table*}

Table~\ref{tab:game-trajectory} illustrates a complete 10-turn simulation using the homogeneous centralized team structure with the RAG-News setting. The team successfully uncovered all four hidden attack cards, achieving victory on the final turn. The table captures each procedure selection, dice roll outcome, and whether the attempt revealed an incident. Additionally, we annotate each turn with whether a retrieval was triggered based on post-failure feedback. Retrieval occurred in 6 out of 10 turns, typically following failed or inconclusive procedure attempts. This highlights how retrieval augmentation is selectively engaged to provide external knowledge support when the team's internal reasoning or roll outcomes are insufficient, improving situational awareness and helping the team recover from earlier setbacks.

\subsection{Experimental Results}

\begin{table}[h!]
    \centering
    \caption{Win rates (\%) and performance gains by team structure in simulated incident response scenarios with and without retrieval augmentation.}
    \label{tab:win-rates}
    \small
    \begin{tabular}{llll}
        \toprule
        \textbf{Team} & \textbf{Base} & \textbf{RAG-Wiki} & \textbf{RAG-News} \\
        \midrule
        Homo. Cen. & 20.0 & 50.0 (+30.0) & \textbf{60.0} (+40.0) \\
        Hetero. Cen. & 30.0 & 43.3 (+13.3) & \textbf{63.3} (+33.3) \\
        Homo. Decen. & 33.3 & 40.0 (+6.7) & \textbf{43.3} (+10.0) \\
        Hetero. Decen. & 26.7 & \textbf{50.0} (+23.3) & \textbf{50.0} (+23.3) \\
        Homo. Hier. & 23.3 & 40.0 (+16.7) & \textbf{43.3} (+20.0) \\
        Hetero. Hier. & 30.0 & 36.7 (+6.7) & \textbf{70.0} (+40.0) \\
        Homo. Arg. & 23.3 & 43.3 (+20.0) & \textbf{46.7} (+23.4) \\
        Hetero. Arg. & 30.0 & 46.7 (+16.7) & \textbf{53.3} (+23.3) \\
        \bottomrule
    \end{tabular}
\end{table}

Table~\ref{tab:win-rates} reports the win rates of all eight team structures across three conditions: base (no retrieval), RAG-Wiki (retrieval from technical webpages), and RAG-News (retrieval from narrative-style incident stories). Across the board, retrieval-augmented teams outperform their base counterparts, often by large margins. Centralized teams benefit significantly from external information, with the homogeneous centralized team improving from 20.0\% to 60.0\% under RAG-News and 50.0\% under RAG-Wiki. Heterogeneous centralized teams show a similar trend, reaching 63.3\% with RAG-News. The impact of retrieval is especially strong for the heterogeneous hierarchical team, which reaches the highest overall performance at 70.0\% with RAG-News, compared to just 30.0\% in the base case. 

Decentralized teams also see meaningful improvements, though the gains are somewhat smaller in homogeneous configurations. Argumentative teams demonstrate notable retrieval benefits despite their lack of centralized control, with the heterogeneous argumentative team improving from 30.0\% to 53.3\% under RAG-News and to 46.7\% under RAG-Wiki. These results suggest that both access to external context and the presence of critical reasoning roles contribute to more successful incident response. Overall, retrieval augmentation enhances team adaptability, improves procedure selection, and reduces the likelihood of overlooking key attack vectors.

\subsection{Ablation Studies}

To understand the sensitivity of AutoBnB-RAG's performance to retrieval design choices, we conduct ablation experiments varying key parameters such as the number of retrieved passages and chunk sizes.

\subsubsection{Effect of Retrieval Numbers}

To assess the impact of retrieval depth, we vary the number of retrieved documents per query (top-$k$) and evaluate performance in the homogeneous centralized team setting. As shown in Table~\ref{tab:top-ks}, both RAG-Wiki and RAG-News settings show relatively stable performance across different values of $k$, with no significant degradation as more documents are retrieved. This suggests that the AutoBnB-RAG framework is reasonably robust to the choice of retrieval depth in the context of this game. In particular, we find that retrieving a small number of documents is often sufficient to provide helpful context for decision-making, while retrieving more documents may offer additional information but also increase the risk of introducing noise.

\begin{table}[h!]
    \centering
    \caption{Win rates (\%) for varying numbers of retrieved documents in the homogeneous centralized team setting.}
    \label{tab:top-ks}
    \small
    \begin{tabular}{lccc}
        \toprule
        \textbf{Team} & \textbf{Top-1} & \textbf{Top-3} & \textbf{Top-5} \\
        \midrule
        RAG-Wiki & 46.7 & 50.0 & 46.7 \\
        RAG-News & 60.0 & 60.0 & 63.3 \\
        \bottomrule
    \end{tabular}
\end{table}

\subsubsection{Effect of Chunk Sizes}

We study the effect of document chunk size by comparing 1,000-character and 5,000-character configurations in the homogeneous centralized team setting. As shown in Table~\ref{tab:chunk-sizes}, larger chunks generally yield higher or comparable win rates, suggesting that preserving more context within each retrieval unit can help agents reason more effectively. This trend is particularly visible in the RAG-Wiki setting, where the win rate improves with longer chunks. However, the difference is less pronounced for RAG-News, indicating that narrative-based retrieval may already provide sufficient coherence even with smaller chunk sizes. Overall, these results suggest that using moderately larger chunks can benefit retrieval-augmented performance, though the optimal size may depend on the nature of the underlying documents.

\begin{table}[h!]
    \centering
    \caption{Win rates (\%) for different document chunk sizes in the homogeneous centralized team setting, using a character-based recursive text splitter.}
    \label{tab:chunk-sizes}
    \small
    \begin{tabular}{lcc}
        \toprule
        \textbf{Team} & \textbf{1k Chars} & \textbf{5k Chars} \\
        \midrule
        RAG-Wiki & 33.3 & 50.0 \\
        RAG-News & 63.3 & 60.0 \\
        \bottomrule
    \end{tabular}
\end{table}

\section{Real-World Simulations}

To assess AutoBnB-RAG's performance in realistic threat scenarios, we simulate real-world cybersecurity incidents drawn from verified news sources. Specifically, we select three high-impact incidents from June 2025~\cite{cmalliance2025june}, representing diverse compromise methods and attacker objectives. Each incident is mapped into a structured Backdoors \& Breaches game, with actual attacker tactics represented by corresponding game cards. Simulations are run using the GPT-4o model with a temperature of 0.7, and incorporate retrieval-augmented generation (RAG) over a curated news corpus. Retrieved content is chunked into 1,000-character windows with overlap to preserve context, and the top three most relevant passages are selected to support each turn. This configuration allows the model to reason fluidly across turns using both in-game dialogue and timely external intelligence.

\subsection{Credential Stuffing on The North Face}

\begin{table*}[h!]
    \centering
    \caption{Turn-by-turn game trajectory from a simulation of the North Face credential stuffing incident using the homogeneous centralized team structure.}
    \label{tab:northface-simulation}
    \small
    \begin{tabular}{clccclc}
        \toprule
        \textbf{Turn} & \textbf{Procedure} & \textbf{Roll} & \textbf{Modifier} & \textbf{Success} & \textbf{Revealed Incident} & \textbf{Retrieval} \\
        \midrule
        1  & User and Entity Behavior Analytics           & 10 & +3 & Yes & Internal Password Spray     & No \\
        2  & SIEM Log Analysis                            & 12 & +3 & Yes & -                          & Yes \\
        3  & Server Analysis                              & 19 & +0 & Yes & Credential Stuffing         & No \\
        4  & Network Threat Hunting - Zeek/RITA Analysis  & 17 & +0 & Yes & HTTPS as Exfil              & No \\
        5  & Endpoint Security Protection Analysis        & 10 & +0 & No  & -                          & Yes \\
        6  & Endpoint Analysis                            & 5  & +0 & No  & -                          & Yes \\
        7  & Endpoint Security Protection Analysis        & 4  & +0 & No  & -                          & Yes \\
        8  & Endpoint Analysis                            & 20 & +0 & Yes & New User Added              & No \\
        \bottomrule
    \end{tabular}
\end{table*}

This simulation models the credential stuffing attack disclosed by The North Face in June 2025, in which customer accounts were accessed using previously breached credentials~\cite{greig2025northface}. Based on the incident details, we mapped the four Backdoors \& Breaches attack stages as follows: \textit{Credential Stuffing} (Initial Compromise), \textit{Internal Password Spray} (Pivot and Escalate), \textit{HTTPS as Exfil} (C2 and Exfiltration), and \textit{New User Added} (Persistence). These cards were explicitly selected to reflect the attacker’s tactics and the observable indicators reported during the breach. The simulation was conducted using a homogeneous centralized team structure, where a team leader coordinated the investigative decisions made by generalist agents. A full turn-by-turn breakdown is provided in Table~\ref{tab:northface-simulation}.

Over the course of eight turns, agents engaged in natural language dialogue to collaboratively propose and refine investigative actions. Early procedures such as \textit{User and Entity Behavior Analytics} successfully revealed internal password spraying, while a failed attempt using \textit{SIEM Log Analysis} triggered a retrieval step that surfaced evidence of credential stuffing. This led to a successful use of \textit{Server Analysis} to uncover the initial compromise. In later turns, the team used \textit{Network Threat Hunting} to detect encrypted exfiltration and employed endpoint-focused procedures to expose persistence mechanisms. Retrieval proved especially valuable after failed or ambiguous rolls, guiding the defenders’ reasoning with contextualized insights. The final success on Turn 8 confirmed unauthorized account creation, completing the full attack path. This case highlights how AutoBnB-RAG's integration of retrieval and structured coordination supports effective detection of complex, multi-stage threats.

\subsection{Roundcube Exploit at Cock.li}

\begin{table*}[h!]
    \centering
    \caption{Turn-by-turn game trajectory from a simulation of the Cock.li Roundcube exploit using the homogeneous centralized team structure.}
    \label{tab:cockli-simulation}
    \small
    \begin{tabular}{clccclc}
        \toprule
        \textbf{Turn} & \textbf{Procedure} & \textbf{Roll} & \textbf{Modifier} & \textbf{Success} & \textbf{Revealed Incident} & \textbf{Retrieval} \\
        \midrule
        1  & Endpoint Security Protection Analysis        & 2  & +3 & No  & -                          & Yes \\
        2  & SIEM Log Analysis                            & 6  & +0 & No  & -                          & Yes \\
        3  & Network Threat Hunting - Zeek/RITA Analysis  & 4  & +0 & No  & -                          & Yes \\
        4  & Server Analysis                              & 12 & +0 & Yes & Web Server Compromise       & No \\
        5  & User and Entity Behavior Analytics           & 8  & +0 & No  & -                          & Yes \\
        6  & Endpoint Analysis                            & 13 & +0 & Yes & Local Privilege Escalation  & No \\
        7  & Network Threat Hunting - Zeek/RITA Analysis  & 19 & +0 & Yes & HTTP as Exfil               & No \\
        8  & Endpoint Security Protection Analysis        & 1  & +3 & No  & -                          & Yes \\
        9  & Endpoint Analysis                            & 7  & +0 & No  & -                          & Yes \\
        10 & Endpoint Security Protection Analysis        & 14 & +3 & Yes & Registry Keys for Persistence & No \\
        \bottomrule
    \end{tabular}
\end{table*}

This simulation modeled the Cock.li data breach~\cite{toulas2025cockli}, where attackers exploited a vulnerability in the Roundcube webmail interface to access over one million user records. The incident was mapped to the following attack stages in the Backdoors \& Breaches framework: \textit{Web Server Compromise} (Initial Compromise), \textit{Local Privilege Escalation} (Pivot and Escalate), \textit{HTTP as Exfil} (C2 and Exfiltration), and \textit{Registry Keys for Persistence} (Persistence). The simulation, summarized in Table~\ref{tab:cockli-simulation}, was carried out using a homogeneous centralized team structure. Initial procedure choices focused on detecting external compromise through log and endpoint analysis, but poor dice rolls delayed early progress and necessitated several strategic pivots supported by retrieval-based insights.

Over 10 turns, the defending agents adapted their investigative approach by leveraging external intelligence to guide their selection of procedures. A key breakthrough came in Turn 4 with a successful \textit{Server Analysis}, which uncovered the web server compromise. In subsequent turns, the team identified privilege escalation through misconfigured endpoints, detected exfiltration via standard HTTP traffic, and ultimately revealed persistence through unauthorized registry key modifications. Throughout the simulation, retrieval augmentation helped the defenders ground their hypotheses in real-world precedent, informing effective and coordinated decisions. Despite initial setbacks, the team successfully uncovered all four attack stages before the turn limit, demonstrating the value of collaborative reasoning, adaptive planning, and retrieval-augmented investigation.

\subsection{Supply Chain Attack on Gluestack}

\begin{table*}[h!]
    \centering
    \caption{Turn-by-turn game trajectory from a simulation of the Gluestack NPM supply chain attack using the homogeneous centralized team structure.}
    \label{tab:gluestack-simulation}
    \small
    \begin{tabular}{clccclc}
        \toprule
        \textbf{Turn} & \textbf{Procedure} & \textbf{Roll} & \textbf{Modifier} & \textbf{Success} & \textbf{Revealed Incident} & \textbf{Retrieval} \\
        \midrule
        1  & SIEM Log Analysis                            & 9  & +3 & Yes & Weaponizing Active Directory         & No  \\
        2  & Endpoint Analysis                            & 2  & +3 & No  & -                                   & Yes \\
        3  & Endpoint Security Protection Analysis        & 17 & +0 & Yes & Malware Injection Into Client Software & No  \\
        4  & Network Threat Hunting - Zeek ...  & 11 & +0 & Yes & Supply Chain Attack                  & No  \\
        5  & Firewall Log Review                          & 8  & +0 & No  & -                                   & Yes \\
        6  & Network Threat Hunting - Zeek ...  & 12 & +0 & Yes & Gmail, Tumblr, Salesforce, Twitter as C2 & No \\
        \bottomrule
    \end{tabular}
\end{table*}

This simulation models the June 2025 supply chain breach of Gluestack's NPM packages~\cite{abrams2025npm}, in which attackers injected remote access trojans into a popular set of React Native libraries. The incident was mapped to the following Backdoors \& Breaches attack stages: \textit{Supply Chain Attack} (Initial Compromise), \textit{Weaponizing Active Directory} (Pivot and Escalate), \textit{Gmail/Tumblr/Salesforce/Twitter as C2} (C2 and Exfiltration), and \textit{Malware Injection Into Client Software} (Persistence). A team of defenders operated under the homogeneous centralized structure, using procedural knowledge, retrieval-supported reasoning, and natural language coordination. Their investigative path, including a series of successful and failed turns, is detailed in Table~\ref{tab:gluestack-simulation}.

Early success in Turn 1 with \textit{SIEM Log Analysis} helped reveal internal Active Directory manipulation, but follow-up endpoint analysis failed to detect signs of initial compromise. Strategic use of retrieval surfaced real-world detection guidance, prompting the team to pursue \textit{Endpoint Security Protection Analysis}, which successfully uncovered the persistence stage. Subsequent network threat hunting revealed both the initial supply chain breach and, after a failed firewall review, the attackers' use of third-party services for covert C2 traffic. The team completed the investigation in six turns, uncovering all four attack stages. This case illustrates how AutoBnB-RAG supports investigative flexibility in stealthy, developer-oriented compromises by blending collaborative reasoning with timely knowledge augmentation.

\section{Conclusion}

This work introduces AutoBnB-RAG, an extension of the AutoBnB framework that integrates retrieval-augmented generation (RAG) into multi-agent incident response simulations. By enabling LLM agents to access external knowledge during collaborative decision-making, AutoBnB-RAG enhances situational awareness, factual grounding, and overall response quality. We evaluate this capability across eight distinct team structures, including newly introduced argumentative configurations designed to foster internal critique and diverse reasoning. Experimental results show that retrieval consistently improves performance, particularly in centralized and hierarchical teams. The use of realistic knowledge sources such as technical documentation and narrative incident reports further demonstrates the adaptability of the RAG approach. To validate the framework in practical contexts, we also simulate real-world breach scenarios, showing that AutoBnB-RAG can effectively reconstruct complex multi-stage attacks through retrieval-informed reasoning. These findings underscore the promise of combining structured multi-agent collaboration with targeted knowledge access to build more capable and resilient AI-driven cyber defense systems.

\bibliographystyle{IEEEtran}
\bibliography{conference}

\begin{thebibliography}{10}
\providecommand{\url}[1]{#1}
\csname url@samestyle\endcsname
\providecommand{\newblock}{\relax}
\providecommand{\bibinfo}[2]{#2}
\providecommand{\BIBentrySTDinterwordspacing}{\spaceskip=0pt\relax}
\providecommand{\BIBentryALTinterwordstretchfactor}{4}
\providecommand{\BIBentryALTinterwordspacing}{\spaceskip=\fontdimen2\font plus
\BIBentryALTinterwordstretchfactor\fontdimen3\font minus \fontdimen4\font\relax}
\providecommand{\BIBforeignlanguage}[2]{{%
\expandafter\ifx\csname l@#1\endcsname\relax
\typeout{** WARNING: IEEEtran.bst: No hyphenation pattern has been}%
\typeout{** loaded for the language `#1'. Using the pattern for}%
\typeout{** the default language instead.}%
\else
\language=\csname l@#1\endcsname
\fi
#2}}
\providecommand{\BIBdecl}{\relax}
\BIBdecl

\bibitem{west1998handbook}
M.~J. West-Brown, D.~Stikvoort, K.-P. Kossakowski, G.~Killcrece, R.~Ruefle, and M.~Zajicek, \emph{Handbook for computer security incident response teams (CSIRTs)}.\hskip 1em plus 0.5em minus 0.4em\relax Carnegie Mellon University, Software Engineering Institute, 1998.

\bibitem{mandia2001incident}
K.~Mandia and C.~Prosise, ``Incident response: investigating computer crime,'' 2001.

\bibitem{kruse2001computer}
W.~G. Kruse~II and J.~G. Heiser, \emph{Computer forensics: incident response essentials}.\hskip 1em plus 0.5em minus 0.4em\relax Pearson Education, 2001.

\bibitem{ahmad2012incident}
A.~Ahmad, J.~Hadgkiss, and A.~B. Ruighaver, ``Incident response teams--challenges in supporting the organisational security function,'' \emph{Computers \& Security}, vol.~31, no.~5, pp. 643--652, 2012.

\bibitem{luttgens2014incident}
J.~T. Luttgens, M.~Pepe, and K.~Mandia, \emph{Incident response \& computer forensics}.\hskip 1em plus 0.5em minus 0.4em\relax McGraw-Hill Education Group, 2014.

\bibitem{naveed2023comprehensive}
H.~Naveed, A.~U. Khan, S.~Qiu, M.~Saqib, S.~Anwar, M.~Usman, N.~Akhtar, N.~Barnes, and A.~Mian, ``A comprehensive overview of large language models,'' \emph{arXiv preprint arXiv:2307.06435}, 2023.

\bibitem{guo2024large}
T.~Guo, X.~Chen, Y.~Wang, R.~Chang, S.~Pei, N.~V. Chawla, O.~Wiest, and X.~Zhang, ``Large language model based multi-agents: A survey of progress and challenges,'' \emph{arXiv preprint arXiv:2402.01680}, 2024.

\bibitem{ni2024mechagents}
B.~Ni and M.~J. Buehler, ``Mechagents: Large language model multi-agent collaborations can solve mechanics problems, generate new data, and integrate knowledge,'' \emph{Extreme Mechanics Letters}, vol.~67, p. 102131, 2024.

\bibitem{wang2024colacare}
Z.~Wang, Y.~Zhu, H.~Zhao, X.~Zheng, T.~Wang, W.~Tang, Y.~Wang, C.~Pan, E.~M. Harrison, J.~Gao \emph{et~al.}, ``Colacare: Enhancing electronic health record modeling through large language model-driven multi-agent collaboration,'' \emph{arXiv preprint arXiv:2410.02551}, 2024.

\bibitem{liu2025econwebarena}
Z.~Liu and Y.~Quan, ``Econwebarena: Benchmarking autonomous agents on economic tasks in realistic web environments,'' \emph{arXiv preprint arXiv:2506.08136}, 2025.

\bibitem{quan2024invagent}
Y.~Quan and Z.~Liu, ``Invagent: A large language model based multi-agent system for inventory management in supply chains,'' \emph{arXiv preprint arXiv:2407.11384}, 2024.

\bibitem{quan2025crmagent}
Y.~Quan, X.~Li, and Y.~Chen, ``Crmagent: A multi-agent llm system for e-commerce crm message template generation,'' \emph{arXiv preprint arXiv:2507.08325}, 2025.

\bibitem{liu2024review}
Z.~Liu, ``A review of advancements and applications of pre-trained language models in cybersecurity,'' in \emph{2024 12th International Symposium on Digital Forensics and Security (ISDFS)}.\hskip 1em plus 0.5em minus 0.4em\relax IEEE, 2024, pp. 1--10.

\bibitem{liu2023secqa}
------, ``Secqa: A concise question-answering dataset for evaluating large language models in computer security,'' \emph{arXiv preprint arXiv:2312.15838}, 2023.

\bibitem{liu2024cyberbench}
Z.~Liu, J.~Shi, and J.~F. Buford, ``Cyberbench: A multi-task benchmark for evaluating large language models in cybersecurity,'' 2024.

\bibitem{tihanyi2024cybermetric}
N.~Tihanyi, M.~A. Ferrag, R.~Jain, T.~Bisztray, and M.~Debbah, ``Cybermetric: A benchmark dataset based on retrieval-augmented generation for evaluating llms in cybersecurity knowledge,'' in \emph{2024 IEEE International Conference on Cyber Security and Resilience (CSR)}.\hskip 1em plus 0.5em minus 0.4em\relax IEEE, 2024, pp. 296--302.

\bibitem{motlagh2024large}
F.~N. Motlagh, M.~Hajizadeh, M.~Majd, P.~Najafi, F.~Cheng, and C.~Meinel, ``Large language models in cybersecurity: State-of-the-art,'' \emph{arXiv preprint arXiv:2402.00891}, 2024.

\bibitem{xu2024large}
H.~Xu, S.~Wang, N.~Li, K.~Wang, Y.~Zhao, K.~Chen, T.~Yu, Y.~Liu, and H.~Wang, ``Large language models for cyber security: A systematic literature review,'' \emph{arXiv preprint arXiv:2405.04760}, 2024.

\bibitem{hays2024employing}
S.~Hays and J.~White, ``Employing llms for incident response planning and review,'' \emph{arXiv preprint arXiv:2403.01271}, 2024.

\bibitem{caviglione2024dawn}
L.~Caviglione, C.~Comito, E.~Coppolillo, D.~Gallo, M.~Guarascio, A.~Liguori, G.~Manco, M.~Minici, S.~Mungari, F.~S. Pisani \emph{et~al.}, ``Dawn of llm4cyber: Current solutions, challenges, and new perspectives in harnessing llms for cybersecurity,'' in \emph{CEUR WORKSHOP PROCEEDINGS}, vol. 3762.\hskip 1em plus 0.5em minus 0.4em\relax CEUR-WS, 2024, pp. 135--140.

\bibitem{backdoorsandbreaches}
{Black Hills Information Security} and {Active Countermeasures}, ``Backdoors \& breaches: An incident response card game,'' \url{https://www.blackhillsinfosec.com/projects/backdoorsandbreaches/}, 2020.

\bibitem{young2021backdoors}
J.~Young and S.~Farshadkhah, ``Backdoors \& breaches: Using a tabletop exercise game to teach cybersecurity incident response,'' in \emph{Proceedings of the EDSIG Conference ISSN}, vol. 2473, 2021, p. 4901.

\bibitem{seiler2025improving}
A.~Seiler and U.~Lechner, ``Improving cyber security incident response: A collaborative tabletop game approach,'' in \emph{IFIP World Conference on Information Security Education}.\hskip 1em plus 0.5em minus 0.4em\relax Springer, 2025, pp. 124--139.

\bibitem{liu2024multi}
Z.~Liu, ``Multi-agent collaboration in incident response with large language models,'' \emph{arXiv preprint arXiv:2412.00652}, 2024.

\bibitem{liu2025autobnb}
------, ``Autobnb: Multi-agent incident response with large language models,'' in \emph{2025 13th International Symposium on Digital Forensics and Security (ISDFS)}.\hskip 1em plus 0.5em minus 0.4em\relax IEEE, 2025, pp. 1--6.

\bibitem{lewis2020retrieval}
P.~Lewis, E.~Perez, A.~Piktus, F.~Petroni, V.~Karpukhin, N.~Goyal, H.~K{\"u}ttler, M.~Lewis, W.-t. Yih, T.~Rockt{\"a}schel \emph{et~al.}, ``Retrieval-augmented generation for knowledge-intensive nlp tasks,'' \emph{Advances in neural information processing systems}, vol.~33, pp. 9459--9474, 2020.

\bibitem{gao2023retrieval}
Y.~Gao, Y.~Xiong, X.~Gao, K.~Jia, J.~Pan, Y.~Bi, Y.~Dai, J.~Sun, M.~Wang, and H.~Wang, ``Retrieval-augmented generation for large language models: A survey,'' \emph{arXiv preprint arXiv:2312.10997}, 2023.

\bibitem{loumachi2025advancing}
F.~Y. Loumachi, M.~C. Ghanem, and M.~A. Ferrag, ``Advancing cyber incident timeline analysis through retrieval-augmented generation and large language models,'' \emph{Computers}, vol.~14, no.~67, pp. 1--42, 2025.

\bibitem{blefari2025cyberrag}
F.~Blefari, C.~Cosentino, F.~A. Pironti, A.~Furfaro, and F.~Marozzo, ``Cyberrag: An agentic rag cyber attack classification and reporting tool,'' \emph{arXiv preprint arXiv:2507.02424}, 2025.

\bibitem{simoni2025morse}
M.~Simoni, A.~Saracino, V.~P, and M.~Conti, ``Morse: Bridging the gap in cybersecurity expertise with retrieval augmented generation,'' in \emph{Proceedings of the 40th ACM/SIGAPP Symposium on Applied Computing}, 2025, pp. 1213--1222.

\bibitem{kurniawan2024cykg}
K.~Kurniawan, E.~Kiesling, and A.~Ekelhart, ``Cykg-rag: Towards knowledge-graph enhanced retrieval augmented generation for cybersecurity,'' 2024.

\bibitem{rahman2024retrieval}
M.~Rahman, K.~O. Piryani, A.~M. Sanchez, S.~Munikoti, L.~De~La~Torre, M.~S. Levin, M.~Akbar, M.~Hossain, M.~Hasan, and M.~Halappanavar, ``Retrieval augmented generation for robust cyber defense,'' Pacific Northwest National Laboratory (PNNL), Richland, WA (United States), Tech. Rep., 2024.

\bibitem{rajapaksha2024rag}
S.~Rajapaksha, R.~Rani, and E.~Karafili, ``A rag-based question-answering solution for cyber-attack investigation and attribution,'' in \emph{European Symposium on Research in Computer Security}.\hskip 1em plus 0.5em minus 0.4em\relax Springer, 2024, pp. 238--256.

\bibitem{zhao2024ontology}
C.~Zhao, G.~Agrawal, T.~Kumarage, Z.~Tan, Y.~Deng, Y.-C. Chen, and H.~Liu, ``Ontology-aware rag for improved question-answering in cybersecurity education,'' \emph{arXiv preprint arXiv:2412.14191}, 2024.

\bibitem{jeon2024rag}
J.-H. Jeon, J.~Koo, and Y.-G. Kim, ``Rag-based cyber threat tracing graph modeling method,'' in \emph{2024 IEEE 23rd International Conference on Trust, Security and Privacy in Computing and Communications (TrustCom)}.\hskip 1em plus 0.5em minus 0.4em\relax IEEE, 2024, pp. 608--615.

\bibitem{wu2023autogen}
Q.~Wu, G.~Bansal, J.~Zhang, Y.~Wu, S.~Zhang, E.~Zhu, B.~Li, L.~Jiang, X.~Zhang, and C.~Wang, ``Autogen: Enabling next-gen llm applications via multi-agent conversation framework,'' \emph{arXiv preprint arXiv:2308.08155}, 2023.

\bibitem{achiam2023gpt}
J.~Achiam, S.~Adler, S.~Agarwal, L.~Ahmad, I.~Akkaya, F.~L. Aleman, D.~Almeida, J.~Altenschmidt, S.~Altman, S.~Anadkat \emph{et~al.}, ``Gpt-4 technical report,'' \emph{arXiv preprint arXiv:2303.08774}, 2023.

\bibitem{langchain}
LangChain, ``Langchain: Framework for developing context-aware language model applications,'' \url{https://www.langchain.com}, 2023.

\bibitem{chroma}
Chroma, ``Chroma: Open-source embedding database and vector search for ai applications,'' \url{https://www.trychroma.com}, 2023.

\bibitem{openai2024embeddings}
OpenAI, ``New embedding models and api updates,'' \url{https://openai.com/index/new-embedding-models-and-api-updates/}, January 2024.

\bibitem{cmalliance2025june}
{CM-Alliance}, ``Major cyber attacks, ransomware attacks and data breaches of june 2025,'' \url{https://www.cm-alliance.com/cybersecurity-blog/major-cyber-attacks-ransomware-attacks-and-data-breaches-of-june-2025}, July 2025.

\bibitem{greig2025northface}
J.~Greig, ``Nearly 3,000 north face website customer accounts breached as retail incidents continue,'' \url{https://therecord.media/north-face-customer-accounts-data-breach-notification}, June 2025.

\bibitem{toulas2025cockli}
B.~Toulas, ``Hacker steals 1 million cock.li user records in webmail data breach,'' \url{https://www.bleepingcomputer.com/news/security/hacker-steals-1-million-cockli-user-records-in-webmail-data-breach/}, June 2025.

\bibitem{abrams2025npm}
L.~Abrams, ``Malware found in npm packages with 1 million weekly downloads,'' \url{https://www.bleepingcomputer.com/news/security/supply-chain-attack-hits-gluestack-npm-packages-with-960k-weekly-downloads/}, June 2025.

\end{thebibliography}
\appendix

The appendix provides supplementary details, including card definitions, prompt templates, and extended simulation settings.
\subsection{Backdoors \& Breaches Cards}
\label{appendix:card-list}

The full set of Backdoors \& Breaches (B\&B)~\cite{backdoorsandbreaches} cards used in our simulations is listed below, organized by attack phase and procedure category:
\begin{itemize}
    \item \textbf{Initial Compromise (13 cards):} Phish, Web Server Compromise, External Cloud Access, Insider Threat, Password Spray, Trusted Relationship, Social Engineering, Bring Your Own (Exploited) Device, Exploitable External Service, Credential Stuffing, Missing HTTP Strict Transport Security (HSTS) Protection, Supply Chain Attack, Physical Access.

    \item \textbf{Pivot and Escalate (12 cards):} Internal Password Spray, Kerberoasting/ASREPRoasting, Broadcast/Multicast Protocol Poisoning, Weaponizing Active Directory, Credential Stuffing, New Service Creation/Modification, Local Privilege Escalation, SMB Weakness, Internal Spearphishing, Access Token Manipulation, Stale Network Address Configurations (SNAC) Attack, Cleartext Passwords in Files.

    \item \textbf{C2 and Exfiltration (7 cards):} HTTP as Exfil, HTTPS as Exfil, DNS as C2, Gmail/Tumblr/Salesforce/Twitter as C2, Domain Fronting as C2, Windows Background Intelligent Transfer Service (BITS), Exfiltration Over Physical Medium.

    \item \textbf{Persistence (14 cards):} Malicious Service, DLL Attacks, Malicious Driver, New User Added, Application Shimming, Malicious Browser Plugins, Logon Scripts, Evil Firmware, Accessibility Features, Event Triggered Malware, Malware Injection Into Client Software, Malicious Email Rules, Windows Service Recovery Actions, Registry Keys for Persistence.

    \item \textbf{Procedure (12 cards):} Security Information and Event Management (SIEM) Log Analysis, Server Analysis, Firewall Log Review, Network Threat Hunting - Zeek/RITA Analysis, Cyber Deception, Endpoint Security Protection Analysis, User and Entity Behavior Analytics (UEBA), Endpoint Analysis, Isolation, Crisis Management, Memory Analysis, Physical Security Review.
\end{itemize}

\subsection{Prompt Template for RAG-News Generation}
\label{prompt-template-news}

The following prompt was used to generate realistic, narrative-style incident response stories for the RAG-News retrieval setting. These stories simulate how a cybersecurity team might investigate and respond to multi-stage attacks using procedure cards from the Backdoors \& Breaches game.

\textbf{Prompt Template:}

Suppose we are writing realistic news-style stories inspired by the Backdoors \& Breaches incident response card game.
In each case, an internal cybersecurity team is responding to a multi-stage attack on their organization.

You will be given a list of attack cards and procedure cards.

Each story should follow this format:

\begin{itemize}[label=-]
    \item Begin the story with a clear and relevant title.
    \item The story should read like a real-world news article or incident report.
    \item Do not include any specific date or timestamp.
    \item The team does not know the attack cards at first.
    \item They must try different procedures, uncover parts of the attack over time, and eventually piece together the full picture.
    \item Include examples of both successful and failed procedures, where applicable.
    \item The team may succeed or fail to stop the attack, but their process must be clear and logical.
    \item Use appropriate cybersecurity reasoning when describing how each procedure helps (or fails) to identify a specific threat.
    \item The attack steps should be revealed one at a time through investigation (not known upfront).
    \item Use plain text. Do not use em dashes (``--'') or emoji. Do not include any extra commentary. Just the story.
\end{itemize}

The attack cards represent the stages of the attack (e.g., initial compromise, escalation, persistence, exfiltration),
and the procedure cards represent detection or investigative methods used by the team.
Some procedures are stronger (e.g., ``Established Procedures'' with a +3 modifier), others are more basic.

Here are the attack cards:

\{attack\_cards\}

Here are the procedure cards:

\{procedure\_cards\}

Write a single, complete news-style story showing how the team investigated and responded to the incident,
gradually uncovering the attack path using the procedure cards. Begin the story with a clear and relevant title.

\subsection{Argumentative Role Definitions}

We introduce a set of argumentative roles designed to enhance team reasoning by encouraging constructive disagreement. These roles mirror their non-argumentative counterparts in expertise but include responsibilities aimed at promoting critical thinking and reducing groupthink during collaborative decision-making.

\subsubsection{Argumentative Team Member}\quad

\textbf{Description:} A generalist role contributing to team strategies while introducing thoughtful disagreements to improve collaborative reasoning.

\textbf{Responsibilities:}
\begin{itemize}
    \item Participate in discussions to analyze the scenario and contribute ideas for the most effective Procedure to use.
    \item Support the team leader in achieving the group's objectives and provide insights from your understanding of the situation.
    \item Respectfully challenge peer suggestions and introduce alternative ideas to stimulate critical thinking and avoid groupthink.
\end{itemize}

\subsubsection{Argumentative Endpoint Security Expert}\quad

\textbf{Description:} Specializes in endpoint protection while promoting rigorous decision-making through constructive argumentation.

\textbf{Responsibilities:}
\begin{itemize}
    \item Analyze endpoints for malware, unauthorized access, or suspicious activities.
    \item Recommend endpoint-specific procedures such as Endpoint Security Protection Analysis and Endpoint Analysis.
    \item Provide detailed insights into securing workstations and detecting endpoint-based attacks.
    \item Raise constructive objections to proposed actions to ensure endpoint-related decisions are thoroughly vetted.
\end{itemize}

\subsubsection{Argumentative Network Traffic Analysis Expert}\quad

\textbf{Description:} Expert in analyzing network threats who enhances team reasoning by deliberately offering alternative interpretations of network data.

\textbf{Responsibilities:}
\begin{itemize}
    \item Examine network traffic logs to detect anomalies and malicious activity.
    \item Recommend procedures such as Network Threat Hunting and Firewall Log Review to monitor and detect threats.
    \item Advocate for tools and strategies to enhance network visibility and security.
    \item Introduce counterpoints to network-related decisions to surface overlooked interpretations or risks.
\end{itemize}

\subsubsection{Argumentative Log and Behavioral Analysis Expert}\quad

\textbf{Description:} Focuses on behavioral and log data while sharpening analysis through respectful disagreement and data reinterpretation.

\textbf{Responsibilities:}
\begin{itemize}
    \item Focus on analyzing logs to detect attack patterns and anomalous user behaviors.
    \item Recommend procedures such as SIEM Log Analysis and User and Entity Behavior Analytics for log-based insights.
    \item Identify and correlate patterns in data that may indicate lateral movement or data exfiltration.
    \item Offer opposing analyses or interpretations of log data to help the team explore multiple investigative angles.
\end{itemize}

\subsubsection{Argumentative Deception and Containment Expert}\quad

\textbf{Description:} Expert in deception and containment who broadens strategy exploration by constructively opposing assumptions.

\textbf{Responsibilities:}
\begin{itemize}
    \item Deploy deception technologies such as honeypots or honeytokens to mislead attackers.
    \item Recommend containment strategies like Isolation to neutralize threats and minimize their impact.
    \item Guide the team in using Cyber Deception effectively to protect high-value assets.
    \item Deliberately question containment timing or strategy to uncover better alternatives or edge cases.
\end{itemize}

\subsubsection{Argumentative Incident Response Expert}\quad

\textbf{Description:} Specialist in active response who enhances resilience by constructively challenging plans and assumptions.

\textbf{Responsibilities:}
\begin{itemize}
    \item Provide expertise on memory analysis and evidence gathering during active incidents.
    \item Suggest procedures such as Memory Analysis and Crisis Management to support effective incident handling.
    \item Guide the team on post-detection actions to contain threats and minimize damage.
    \item Intentionally probe the robustness of proposed incident handling steps to uncover gaps or oversights.
\end{itemize}

\subsection{Simulation Turn Sequence}

The following sequence defines the structure of a full simulation game within the AutoBnB-RAG framework, adapted from the rules of the Backdoors \& Breaches game.

\begin{enumerate}
    \item \textbf{Set the Scenario}
    \begin{itemize}
        \item Select one card for each of the four attack stages: Initial Compromise, Pivot and Escalate, C2 and Exfil, and Persistence.
        \item Craft a detailed initial scenario description based on the selected Initial Compromise card. Provide sufficient context for Defenders to understand the situation without revealing card names or direct clues.
    \end{itemize}
    
    \item \textbf{Introduce the Defenders to the Procedure Cards}
    \begin{itemize}
        \item Explain the difference between Established Procedures (with a +3 modifier) and Other Procedures (with a +0 modifier).
        \item Present the full list of Procedure cards and identify which are classified as Established.
    \end{itemize}
    
    \item \textbf{Start Each Turn (Turn 1 to Turn 10)}
    \begin{itemize}
        \item Announce the current turn number.
        \item Prompt the Defenders to discuss and collaboratively select one Procedure card to use.
    \end{itemize}
    
    \item \textbf{Defenders' Procedure Attempt}
    \begin{itemize}
        \item Roll a 20-sided die to resolve the Procedure attempt.
        \item Apply the appropriate modifier:
        \begin{itemize}
            \item Established Procedure: +3 modifier
            \item Other Procedure: +0 modifier
        \end{itemize}
        \item Determine the result:
        \begin{itemize}
            \item Final roll of 11 or higher: success
            \item Final roll of 10 or lower: failure
        \end{itemize}
    \end{itemize}
    
    \item \textbf{Respond to Success or Failure}
    \begin{itemize}
        \item On Success: If the Procedure matches a detection method for any unrevealed attack card, reveal one such card to the Defenders.
        \item On Failure: Notify the Defenders that no attack card was revealed.
    \end{itemize}
    
    \item[5+)] \textbf{Post-Attempt Retrieval}
    \begin{itemize}
        \item After resolving the Procedure attempt, the Incident Captain may issue a retrieval query if the team needs clarification or context.
        \item Retrieved information is shared with all agents to assist in future decisions.
    \end{itemize}
    
    \item \textbf{End Game}
    \begin{itemize}
        \item Victory: All four attack cards are revealed within 10 turns.
        \item Loss: Fewer than four attack cards are revealed after 10 turns.
        \item Save a summary of the simulation, including major events and final outcome.
    \end{itemize}
\end{enumerate}

\end{document}